\documentclass[10pt,twocolumn,letterpaper]{article}
\PassOptionsToPackage{table}{xcolor}

\usepackage[pagenumbers]{cvpr} 
\usepackage{threeparttable} 
\usepackage{stfloats}

\definecolor{cvprblue}{rgb}{0.21,0.49,0.74}
\usepackage[pagebackref,breaklinks,colorlinks,allcolors=cvprblue]{hyperref}

\def\paperID{25} 
\def\confName{CVPR}
\def\confYear{2025}

\title{Team of One: Cracking Complex Video QA with Model Synergy}

\author{
Jun Xie$^{1,*}$, Zhaoran Zhao$^{1,*,\dag}$, Xiongjun Guan$^{2}$, Yingjian Zhu$^{3,4}$, \\
Hongzhu Yi$^{3}$,Xinming Wang$^{4,5}$, Feng Chen$^{1}$, Zhepeng Wang$^{1}$\\
$^1$Lenovo research \qquad  $^2$ Tsinghua University \\
$^3$School of Artificial Intelligence, University of Chinese Academy of Sciences (UCAS) \\
$^4$Institute of Automation, Chinese Academy of Sciences(CAS) 
$^5$Zhongguancun Academy\\
{\tt\small \{xiejun, zhaozr3, chenfeng13, wangzpb\}@lenovo.com,} \\ 
{\tt\small \{zhuyingjian24,yihongzhu23\}@mails.ucas.ac.cn,}\\
{\tt\small wangxinming2024@ia.ac.cn,} 
{\tt\small gxj21@mails.tsinghua.edu.cn}
}

\begin{document}
\maketitle

\begin{NoHyper}
\def\thefootnote{}\footnotetext{* These authors contributed equally.}
\end{NoHyper}
\begin{NoHyper}
\def\thefootnote{}\footnotetext{$\dag$ Corresponding author.}
\end{NoHyper}

\begin{abstract}
We propose a novel framework for open-ended video question answering that enhances reasoning depth and robustness in complex real-world scenarios, as benchmarked on the CVRR-ES dataset. Existing Video-Large Multimodal Models (Video-LMMs) often exhibit limited contextual understanding, weak temporal modeling, and poor generalization to ambiguous or compositional queries. To address these challenges, we introduce a prompting-and-response integration mechanism that coordinates multiple heterogeneous Video-Language Models (VLMs) via structured chains of thought, each tailored to distinct reasoning pathways. An external Large Language Model (LLM) serves as an evaluator and integrator, selecting and fusing the most reliable responses. Extensive experiments demonstrate that our method significantly outperforms existing baselines across all evaluation metrics, showcasing superior generalization and robustness. Our approach offers a lightweight, extensible strategy for advancing multimodal reasoning without requiring model retraining, setting a strong foundation for future Video-LMM development.

\end{abstract}    
\section{Introduction}
\label{sec:intro}

CVRR-ES is a comprehensive, open-ended video question-answering benchmark specifically designed to evaluate the reasoning depth and robustness of Video-Large Multimodal Models (Video-LMMs) in complex, real-world scenarios~\cite{khattak2024good}. It spans 11 diverse and world-centric video dimensions—including social interactions, emotional expressions, and anomalous physical events—featuring videos with high contextual and temporal dependency. The benchmark comprises carefully curated open-ended textual queries that challenge models to interpret nuanced visual and temporal information. By focusing on scenarios that require multi-step reasoning and resilience to varied linguistic inputs, CVRR-ES serves as a rigorous diagnostic suite for assessing the real-world applicability of Video-LMMs. It provides a critical foundation for benchmarking progress in building human-aligned, robust, and context-aware multimodal systems.

Existing methods~\cite{Video-llava,li2024llama,zhang2023video} for Video-Large Multi-modal Models (Video-LMMs) primarily focus on enhancing architectural components, expanding task coverage, or improving video length compatibility, yet they exhibit notable limitations in reasoning and robustness. While recent models like Video-LLaVA~\cite{zhang2023video} and VideoChatGPT~\cite{maaz2023video} have made strides in integrating vision and language through connector modules and training objectives, they largely overlook the need for precise reasoning over complex, real-world video content and remain sensitive to the quality and structure of user queries. 
These methods struggle with contextual dependency, nuanced understanding, and compositional reasoning required for real-world video scenarios. 
Moreover, training-free prompting~\cite{wang2022self,wei2022chain}—a lightweight and generalizable strategy successfully adopted in NLP—remains underexplored in the Video-LMM domain, limiting the adaptability of these models during inference for tasks demanding deeper comprehension and robust response generation.

\begin{figure*}[!t]
	\centering
	\includegraphics[width=.95\linewidth]{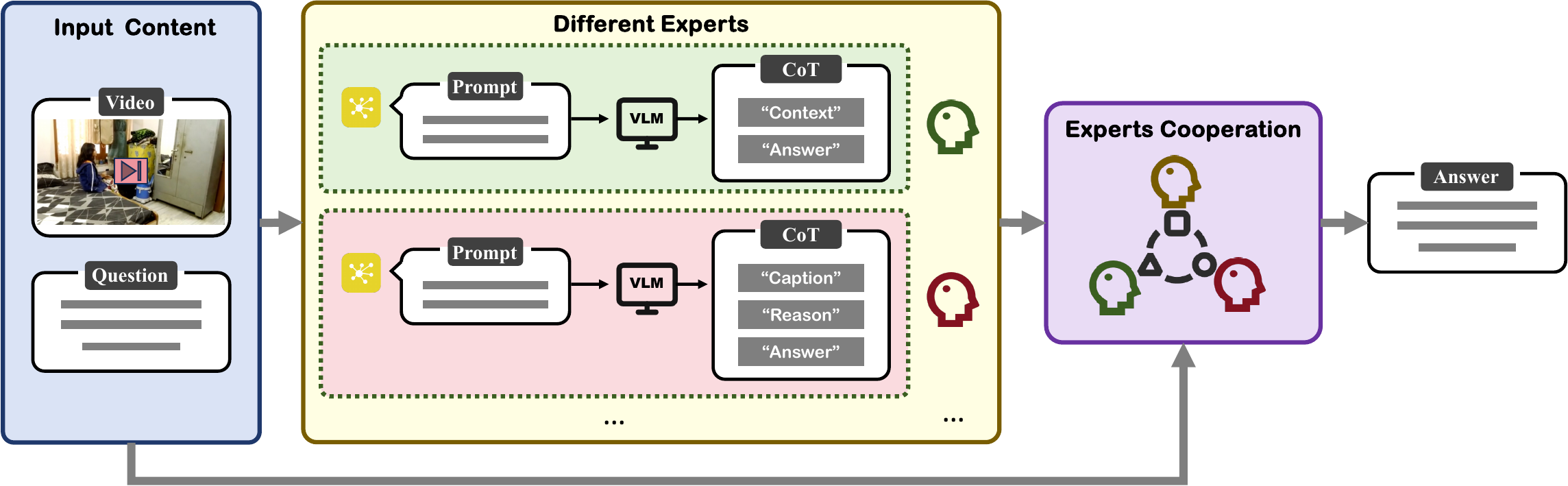}
	\vspace{1mm}
	\caption{The framework of our proposed algorithm. The specific text content is omitted in the flowchart and reported in the main text.}
	\label{fig:framework}
\end{figure*}

To enhance the reasoning depth and robustness of models in complex real-world scenarios, we propose a novel open-ended video question answering framework featuring a prompt-response ensemble mechanism. This mechanism leverages structured Chain-of-Thought prompting to coordinate multiple heterogeneous video-language models (VLMs), with each reasoning chain tailored to explore a distinct inference pathway, enabling multi-perspective and multi-level understanding of complex video content.
In addition, we incorporate an external Multimodal Large Language Model (MLLM) as an evaluator and aggregator to assess multiple candidate responses and integrate the most reliable one as the final output.
Extensive experimental results demonstrate that our method significantly outperforms existing state-of-the-art baselines across all evaluation metrics, exhibiting superior generalization and robustness. As a lightweight and scalable approach that requires no retraining of underlying models, our framework effectively advances multimodal reasoning capabilities and lays a solid foundation for building robust Video-LMMs capable of real-world understanding.

\section{Method}
\label{sec:method}

\begin{table}[h!]
\centering
\caption{Benchmark categories and their corresponding category numbers.}
\begin{tabular}{p{3.5cm}|c}
\hline
\textbf{Benchmark Category} & \textbf{Category Number} \\
\hline
Multiple Actions in single video. & 1 \\
\hline
Fine-grained action understanding. & 2 \\
\hline
Partial actions. & 3 \\
\hline
Time order understanding. & 4 \\
\hline
Non-existent actions with existent scene. & 5 \\
\hline
Non-existent actions with non-existent scene. & 6 \\
\hline
Continuity and Object instance Count. & 7 \\
\hline
Unusual and Physically Anomalous activities. & 8 \\
\hline
Interpretation of social context. & 9 \\
\hline
Understanding of emotional context. & 10 \\
\hline
Interpretation of visual context. & 11 \\
\hline
\end{tabular}

\label{tab:benchmark_categories}
\end{table}

Our proposed framework (as illustrated in Figure~\ref{fig:framework}) is designed to systematically address a series of core challenges encountered by current Video-Language Multimodal Models in complex video reasoning tasks. These challenges include: a tendency to produce overconfident answers when faced with ambiguous or incomplete queries, strong prior assumptions regarding action completion, limited generalization to out-of-distribution (OOD) samples, weak understanding of temporal event progression, and insufficient modeling of high-level semantics such as emotion and social context.

To tackle these issues, we introduce a general and highly extensible prompt-and-response integration mechanism. Specifically, for each input instance—consisting of a video and its corresponding natural language question—we construct diverse prompting strategies, design multi-path information flows, and incorporate structured chains of thought (CoT) ~\cite{wei2022chain} to guide multiple heterogeneous Video-Language Models in reasoning from different semantic perspectives. Each chain of thought is carefully designed to direct the model’s attention to specific visual cues or semantic elements, thereby generating logically diverse and complementary candidate responses.

On top of this, we incorporate a Multimodal Large Language Model (MLLM) as both an evaluator and integrator. This module performs semantic consistency evaluation across candidate responses based on the original question and video context, selecting the final answer that best aligns with both the intended semantics and visual evidence. This integration mechanism effectively mitigates the reasoning bias of individual models and significantly enhances the robustness and generalization ability of the system in complex video question-answering scenarios. Furthermore, the framework is readily extensible, allowing future integration of additional expert models.








\subsection{Model Selection and Prompt Design}

We first identify suitable large-scale Video-Language Models for our task. The selection criteria include: strong multimodal comprehension for video understanding, robust logical reasoning for complex queries, and proficient language generation capabilities. Considering both performance and computational cost, we adopt the Gemini model series~\cite{Google_Gemini_2_5_Pro_2025} as our primary implementation backbone.

For prompt design, we construct four distinct prompting pathways tailored to different reasoning demands:

\begin{enumerate}
    
    \item Capturing the contextual semantics of the query from multiple perspectives to avoid single-view bias;
    \item Emphasizing whether the described action truly occurs in the video, in order to mitigate hallucination issues;
    \item Modeling the temporal sequence and causal relations of events to enhance fine-grained temporal reasoning;
    \item Adopting a question-driven prompting approach that encourages the model to generate answers by focusing explicitly on the query before processing the video.
\end{enumerate}

This design effectively directs the model's attention toward semantic details and relevant visual cues, facilitating the generation of diverse and logically sound responses in complex multimodal reasoning scenarios.

\subsection{Model Integration}

\begin{table*}[!htbp]
  \centering
  \begin{threeparttable}
    \setlength{\tabcolsep}{12pt}
    \caption{Performance comparison on CVRR-ES between the proposed method and public baselines.}
    \label{tab:leaderboard}
    \small
    \begin{tabular}{p{2.9cm} | *{11}{p{0.33cm}} p{0.6cm}}
      \toprule
      Benchmark Category & 1 & 2 & 3 & 4 & 5 & 6 & 7 & 8 & 9 & 10 & 11 & Avg. \\
      \hline
      Video-LLaMA-2 & 16.98 & 29.57 & 24.76 & 16.45 & 10.14 & 13.19 & 28.25 & 18.95 & 25.00 & 21.92 & 32.60 & 21.62 \\
      VideoChat & 23.90 & 33.48 & 33.01 & 31.58 & 15.22 & 14.58 & 24.29 & 18.42 & 31.07 & 23.63 & 34.43 & 25.78 \\
      Video-ChatGPT & 27.67 & 26.96 & 22.82 & 27.63 & 23.19 & 17.36 & 28.41 & 18.95 & 32.50 & 21.23 & 27.84 & 24.96 \\
      Video-LLaVA & 15.72 & 25.22 & 13.59 & 21.05 & 5.07 & 3.47 & 21.47 & 15.79 & 18.93 & 15.07 & 19.78 & 15.92 \\
      MovieChat & 12.58 & 23.48 & 21.36 & 16.45 & 5.07 & 11.81 & 19.77 & 17.89 & 17.14 & 13.70 & 21.25 & 16.41 \\
      LLaMA-VID & 17.92 & 26.09 & 14.56 & 19.74 & 2.90 & 6.94 & 24.86 & 16.32 & 13.93 & 14.73 & 23.08 & 16.46 \\
      TimeChat & 28.30 & 39.13 & 49.51 & 34.21 & 23.19 & 13.89 & 34.46 & 27.37 & 39.29 & 27.40 & 45.05 & 32.89 \\
      Gemini-V Pro & 43.08 & 51.61 & 67.48 & 45.39 & 57.25 & 49.64 & 36.16 & 60.00 & 64.29 & 47.26 & 63.00 & 53.20 \\
      GPT-4V & 57.55 & 77.39 & 73.79 & 57.89 & 71.01 & 75.00 & 62.71 & 74.74 & 79.64 & 66.44 & 82.42 & 70.78 \\
      \hline
      Human & 93.40 & 95.65 & 98.54 & 97.37 & 97.10 & 100.00 & 96.49 & 96.84 & 97.51 & 95.55 & 94.87 & 96.67 \\
            \rowcolor{gray!10} Ours & 84.12 & 86.75 & 91.29 & 79.00 & 96.02 & 92.14 & 76.35 & 91.22 & 94.43 & 87.04 & 90.09 & 88.04 \\
      \bottomrule
    \end{tabular}
    \begin{tablenotes}
      \footnotesize
      \item[$\dag$] Reported from \cite{khattak2024good}.
    \end{tablenotes}
  \end{threeparttable}
\end{table*}

To fully leverage the complementary strengths of different expert models in multimodal reasoning, we propose a response integration mechanism based on a Multimodal Large Language Model (MLLM). Specifically, we feed the MLLM with the original question, video content, and multiple candidate answers generated by diverse expert pathways. Each pathway is explicitly annotated with its unique reasoning characteristics—such as temporal structure modeling, assessment of action plausibility, or question-driven analytical strategies—to help the MLLM understand the reasoning biases and advantages of each expert.

\begin{table}[h!]
\centering
\caption{Leaderboard rankings on validation set}
\begin{tabular}{c|l|c}
\hline
Rank & Participant Teams & Acc (↑) \\
\hline
1 &  FRI & 0.53 \\
2 & Host\_6403\_Team & 0.63 \\
3 & NJUST\_\_KMG & 0.85 \\
\rowcolor{gray!20} 4 & PCIEgogogo (Ours) & 0.88 \\
5 & TeamCVPR2025 & 0.92 \\
\hline
\end{tabular}

\label{tab:leaderboard_simple}
\end{table}

During integration, the MLLM first performs semantic parsing of the question to infer the most suitable reasoning pathway. It then jointly considers the video and question context to evaluate and compare the candidate answers, selecting and fusing the most relevant and reliable information. This mechanism not only significantly enhances the system’s robustness and generalization ability in complex reasoning tasks but also offers strong scalability for integrating additional heterogeneous expert models in the future.

\section{Experiments} \label{sec:experiments}


Table ~\ref{tab:leaderboard} presents the performance of our method on the validation set. For clarity, we index the evaluation dimensions numerically, with the corresponding category-to-index mapping detailed in Table ~\ref{tab:benchmark_categories}. Our method consistently outperforms all existing public baselines across all metrics, demonstrating superior generalization, temporal reasoning capability, and robustness in complex video scenarios.
Table ~\ref{tab:leaderboard_simple} further compares our method against non-public baselines on the validation set. On the latest released test set, our approach ranks third achieving an average accuracy of 0.75 (Jun 5, 2025 8:00:59 PM CST).
It is worth noting that there exists a noticeable performance gap between the training and validation sets for our method. This discrepancy may stem from differences in data distribution or annotation strategies, which present additional challenges to model generalization.



\section{Conclusion} \label{sec:conclusion}

In this paper, we present a general and extensible prompting-and-response integration framework for improving the reasoning capabilities of Video-Large Multimodal Models on the CVRR-ES benchmark. By leveraging diverse prompt strategies, multi-path reasoning chains, and an MLLM-based evaluator for response integration, our method systematically addresses key limitations in existing models, including overconfidence on ambiguous queries, weak temporal and semantic modeling, and poor generalization. Experimental results show that our approach consistently surpasses public baselines in both validation and test scenarios, despite using only locally evaluated results. Our findings highlight the effectiveness of combining lightweight prompting techniques with expert model coordination, providing a scalable and training-free solution for robust multimodal video reasoning in open-ended, real-world tasks.

{
    \small
    \bibliographystyle{ieeenat_fullname}
    \bibliography{main}
}



\end{document}